\documentclass[conference,a4paper]{IEEEtran}
\IEEEoverridecommandlockouts

\usepackage[hidelinks]{hyperref}
\usepackage[cmex10]{amsmath}
\usepackage{amssymb}
\usepackage{dblfloatfix}

\usepackage{graphicx}
\graphicspath{{.}}

\usepackage{booktabs}
\usepackage{cite}
\usepackage{orcidlink}
\usepackage{tikz}
\usetikzlibrary{arrows.meta, positioning, calc, backgrounds, fit}

\newcommand\copyrighttext{%
  \footnotesize \textcopyright~\the\year{} IEEE. Personal use of this material is permitted.
  Permission from IEEE must be obtained for all other uses, in any current or future media,
  including reprinting/republishing this material for advertising or promotional purposes,
  creating new collective works, for resale or redistribution to servers or lists, or reuse
  of any copyrighted component of this work in other works.}
\newcommand\copyrightnotice{%
\begin{tikzpicture}[remember picture,overlay]
\node[anchor=south,yshift=10pt] at (current page.south)
{\parbox{\dimexpr\textwidth-2\fboxsep\relax}{\copyrighttext}};
\end{tikzpicture}%
}

\begin{document}

\title{T-SAR-JEPA: Self-Supervised Temporal Anomaly Detection in SAR Amplitude Stacks via Latent Prediction}

\author{
	\IEEEauthorblockN{Kerod Woldesenbet\textsuperscript{*}\orcidlink{0009-0000-2335-2429}}
	\IEEEauthorblockA{\textit{Independent Researcher}\\
		kerod5858@gmail.com}
	\and
	\IEEEauthorblockN{Abem Woldesenbet\textsuperscript{*}\orcidlink{0009-0002-6495-0091}}
	\IEEEauthorblockA{\textit{Dakota State University}\\
		abem.woldesenbet@trojans.dsu.edu}
	\thanks{\textsuperscript{*}These authors contributed equally to this work.}
}

\maketitle
\copyrightnotice
\begin{abstract}
We present T-SAR-JEPA, a self-supervised framework for temporal anomaly detection in SAR amplitude stacks via latent prediction. A ViT-Base/16 encoder from SAR-JEPA~\cite{li2024sarjepa} is domain-adapted on 39{,}300 Capella patches using local masked reconstruction~\cite{chen2025lomar} with gradient feature prediction. A temporal transformer with sinusoidal time encoding forecasts future latent states from $K{=}7$ acquisitions, with progressive unfreezing substantially reducing validation loss. The model operates on amplitude alone; InSAR coherence serves exclusively as independent pseudo-ground-truth. On the DFC~2026 dataset~\cite{dfc2026} (300 time-series, three AOIs), T-SAR-JEPA achieves ROC-AUC of \textbf{77.0}\% on the Hawaii eruption window, outperforming RX, PaDiM, Linear AR, and LSTM baselines (${\sim}50$\%). Spatial coherence of \textbf{99.9}\% ($p < 0.001$, permutation test) confirms structured detections. Code: \url{https://github.com/TerraLatent/t-sar-jepa}.
\end{abstract}

\begin{IEEEkeywords}
SAR, self-supervised learning, JEPA, anomaly detection, temporal analysis, change detection
\end{IEEEkeywords}

% ============================================================
\section{Introduction}
% ============================================================

Synthetic aperture radar (SAR) enables all-weather, day-night surface monitoring~\cite{bamler1998insar}. Traditional temporal change detection relies on InSAR~\cite{rosen2000insar_review, ferretti2001insar}, requiring co-registration, phase unwrapping, and atmospheric correction~\cite{berardino2002sbas}, limiting scalability to dense commercial stacks. Self-supervised learning (SSL) has been applied to SAR despeckling~\cite{sar2sar}, spatial anomaly detection~\cite{sariad2025}, and bitemporal change detection~\cite{skycap2025}, but these methods do not model multi-temporal dynamics from long acquisition stacks. Joint Embedding Predictive Architectures (JEPA)~\cite{lecun2022jepa, assran2023ijepa} predict representations rather than pixels, well-suited to SAR where speckle corrupts pixel-level targets. Li et al.~\cite{li2024sarjepa} demonstrated SAR-JEPA for automatic target recognition, training a ViT encoder to predict multi-scale gradient features of masked patches from visible context via local masked reconstruction~\cite{chen2025lomar}.

Most deep learning approaches to SAR change detection operate on amplitude alone~\cite{sariad2025, skycap2025}. Recent multitemporal SSL methods reconstruct masked pixels across time~\cite{labatie2026maestro}, while JEPA-based retrieval~\cite{choudhury2025rejepa} and modality-agnostic foundation models~\cite{xiong2024dofa} provide complementary directions. We adopt the amplitude-only paradigm but use InSAR coherence as \emph{independent validation}: our model discovers temporal anomalies from amplitude, and coherence independently confirms these correspond to physical surface change.

\begin{figure*}[t]
\centering
    \includegraphics[width=\textwidth]{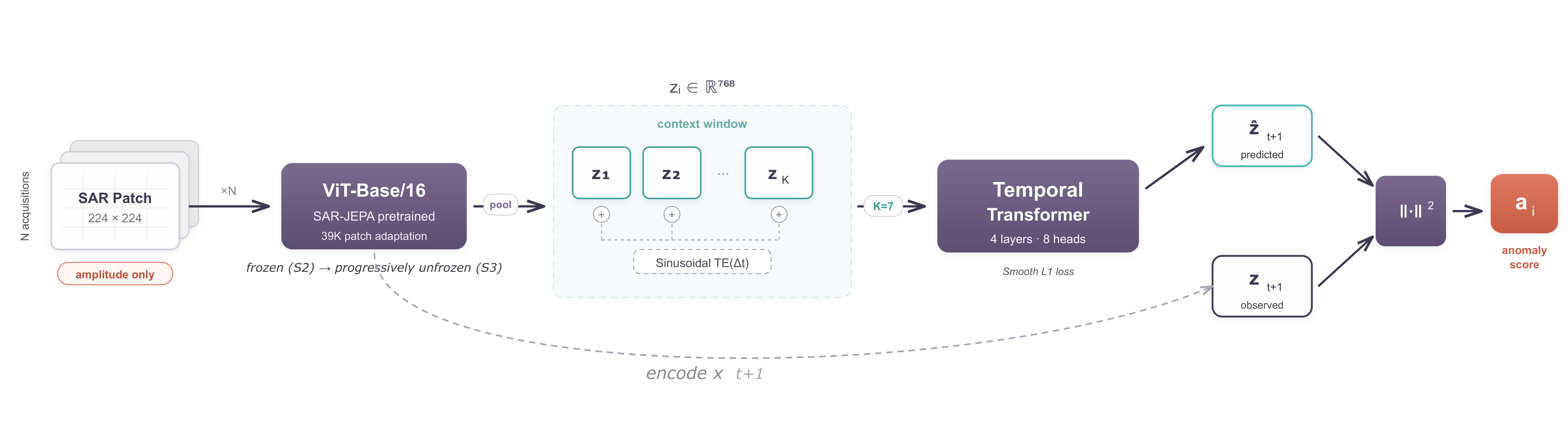}
\caption{T-SAR-JEPA inference dataflow. Single-channel SAR amplitude patches ($1{\times}224{\times}224$) from $N$ acquisitions are independently encoded by a ViT-Base/16 (SAR-JEPA pretrained, domain-adapted on 39K Capella patches) into $z_i \in \mathbb{R}^{768}$. A context window of $K{=}7$ embeddings, augmented with sinusoidal time encodings $\text{TE}(\Delta t)$, feeds a 4-layer temporal transformer that predicts $\hat{z}_{t+1}$. The anomaly score $a_i = \|\hat{z}_i - z_i\|_2$ measures prediction error against the observed embedding. The encoder is frozen during Stage~2 and progressively unfrozen in Stage~3. InSAR coherence serves as independent validation only --- it is never seen by the model.}
\label{fig:architecture}
\end{figure*}

We introduce T-SAR-JEPA (Fig.~\ref{fig:architecture}), which: (1)~domain-adapts a pretrained SAR-JEPA encoder on 39K Capella patches; (2)~trains a temporal transformer with sinusoidal time encodings on frozen latent sequences; (3)~applies progressive unfreezing for end-to-end refinement; and (4)~scores anomalies as L2 prediction errors validated against coherence-derived pseudo-GT across three AOIs from DFC~2026~\cite{dfc2026}.

\textbf{Contributions:} (1)~First JEPA-based temporal SAR anomaly detector on amplitude alone, with InSAR coherence as independent validation; (2)~Sinusoidal time encoding outperforms learnable alternatives with ablation across three encoding types; (3)~Progressive unfreezing yields substantial validation loss reduction over frozen-encoder training; (4)~Evaluation against four baselines with ROC/PR metrics and permutation-tested spatial coherence; (5)~Public code release.

% ============================================================
\section{Method}
% ============================================================

T-SAR-JEPA operates in three stages: encoder domain adaptation, temporal predictor training, and end-to-end fine-tuning.

\subsection{Stage 1: Encoder Domain Adaptation}

We build on SAR-JEPA~\cite{li2024sarjepa}, a ViT-Base/16~\cite{dosovitskiy2021vit} pretrained with local masked reconstruction~\cite{chen2025lomar} to predict multi-scale gradient features (edge and texture descriptors) of masked regions from visible context on SAR ATR datasets. The encoder uses a single-channel projection \texttt{Conv2d(1, 768, 16, 16)}, accepting log-dB amplitude normalized via P2/P98 (2nd/98th percentile) clipping to $[0,1]$. We fine-tune on 39{,}300 Capella patches across all three AOIs using the same objective, bridging the distribution gap between pretraining data (MSTAR/SAR-ACD) and the target Capella constellation. The output is a 768-dimensional representation per $224{\times}224$ patch.

\subsection{Stage 2: Temporal Predictor}

The adapted encoder is frozen and used to extract $z_i \in \mathbb{R}^{768}$ per acquisition. A 4-layer transformer~\cite{vaswani2017attention} (23.2M params, 768 hidden, 8 heads, 2048 MLP, dropout 0.1) forecasts $\hat{z}_i$ from the $K{=}7$ preceding representations. Capella acquisitions are irregularly spaced (typically 2--5 days apart), so integer positional indices discard the physical elapsed time between frames. We adopt fixed sinusoidal encodings as a continuous function of $\Delta t$: a geometric progression of frequencies gives a multi-scale periodic basis across timescales without per-frequency calibration, remains well-conditioned in the low-data regime (300 sequences) where learnable encodings are under-determined, and extrapolates in closed form to unseen cadences at inference. We embed the elapsed time $\Delta t$ (in days) as
\begin{equation}
    \text{TE}(\Delta t) = [\sin(\omega_k \Delta t),\, \cos(\omega_k \Delta t)]_{k=1}^{384}
\end{equation}
where $\omega_k = 1/10000^{2k/d}$ for $k = 1, \ldots, 384$ ($d{=}768$), following the standard geometric progression~\cite{vaswani2017attention} that spans periods from ${\sim}1$~day to ${\sim}10^4$~days. At a typical 3-day cadence, $K{=}7$ corresponds to ${\sim}$3 weeks of lookback. We ablate against CTLPE~\cite{ctlpe2024} and linear learnable encodings (Sec.~\ref{sec:ablations}). Training minimizes Smooth L1 loss (robust to speckle-induced outliers):
\begin{equation}
    \mathcal{L}_{\text{temporal}} = \frac{1}{N{-}K} \sum_{i=K+1}^{N} \text{SmoothL1}(\hat{z}_i,\, z_i),
\end{equation}
where $z_i \in \mathbb{R}^{768}$ is the encoder embedding of acquisition~$i$, $\hat{z}_i$ is its prediction from the preceding $K$ embeddings, and $N$ is the sequence length. The sum starts at $i{=}K{+}1$ since the first $K$ acquisitions lack a full context window.

\subsection{Stage 3: Progressive Unfreezing}

After Stage~2 convergence, the encoder is progressively unfrozen with differential learning rates. Phase~A unfreezes the last 4 encoder blocks (encoder $\text{lr}=10^{-5}$, predictor $\text{lr}=10^{-4}$, 30 epochs). Phase~B unfreezes all layers with halved rates (20 epochs). This aligns encoder representations with the temporal prediction task while preserving pretrained low-level features~\cite{howard2018ulmfit}.

\subsection{Anomaly Scoring}

Anomaly scores are L2 distances between predicted and actual representations:
\begin{equation}
    a_i = \| \hat{z}_i - z_i \|_2
\end{equation}
We use L2 rather than the training loss (Smooth L1) because it preserves outlier magnitude, which is the detection signal. Stable surfaces produce low $a_i$; surface changes break temporal predictability, elevating $a_i$. Detections use a percentile threshold swept over P70--P90 of the per-AOI score distribution.

% ============================================================
\section{Experiments}
% ============================================================

\subsection{Dataset and Preprocessing}

We use the DFC~2026 Capella Space dataset~\cite{dfc2026, capella2023} across three AOIs: \textbf{Hawaii/Kilauea} (186 collects, volcanic terrain with Dec~2024 eruption), \textbf{Los Angeles} (115 collects, semi-arid urban/forest), and \textbf{Pilbara/W.~Australia} (92 collects, arid mining, mixed satellites C09/C10/C14). From each GEO image we extract a $10{\times}10$ grid of $224{\times}224$ patches from the center footprint, yielding 39{,}300 patches and 300 temporal sequences. Amplitude is converted to log~dB, P2/P98 clipped, and normalized to $[0,1]$.

\textbf{Pseudo-GT.} InSAR coherence maps from 19 SLC pairs around the Kilauea eruption window (Dec 2024--Jan 2025) serve as independent pseudo-ground-truth. Coherence drops below 0.2 define ``change'' events (99 positive out of 1{,}900 pairs, 5.2\% prevalence). This coherence data is \emph{never seen by the model}.

\subsection{Baselines}

All baselines operate on the same frozen Stage~1 ViT encoder features. We note that baselines do not receive Stage~3 progressive unfreezing; the ablation in Table~\ref{tab:ablation} shows frozen T-SAR-JEPA (val loss $2.0 \times 10^{-3}$) as a fairer temporal comparison, while unfreezing (val loss $0.04 \times 10^{-3}$) demonstrates the additional gain from encoder--predictor co-adaptation. Baselines:
\textbf{RX}~\cite{reed1990rx} (Mahalanobis-distance anomaly detector, non-temporal),
\textbf{PaDiM}~\cite{defard2021padim, padim_ace} (per-location Gaussian modeling, non-temporal),
\textbf{Linear AR} (autoregressive order $K{=}7$, temporal), and
\textbf{LSTM} (2-layer, 512 hidden, temporal).
RX and PaDiM serve as non-temporal reference points; Linear AR and LSTM are the direct temporal comparisons.

\subsection{Implementation Details}

\begin{table}[t]
    \centering
    \caption{Training configuration for each stage.}
    \label{tab:impl}
    \small
    \setlength{\tabcolsep}{3pt}
    \begin{tabular}{lccc}
        \toprule
        & \textbf{Stage 1} & \textbf{Stage 2} & \textbf{Stage 3} \\
        \midrule
        Optimizer & \multicolumn{3}{c}{AdamW, cosine decay} \\
        Learning rate & $10^{-4}$ & $10^{-4}$ & $10^{-5}$/$10^{-4}$ \\
        Weight decay & 0.05 & 0.01 & 0.01 \\
        Batch size & 256 & 64 & 64 \\
        Epochs & 50 & 150 & 30+20 \\
        Encoder & fine-tune & frozen & progressive \\
        \bottomrule
    \end{tabular}
\end{table}

Table~\ref{tab:impl} summarizes training hyperparameters. Stage~2 uses an 80/20 spatial split (disjoint grid locations) to prevent temporal leakage; Stage~3 Phase~A unfreezes the last 4 blocks, Phase~B all layers with halved rates and gradient clipping at 1.0. All training on a single NVIDIA H200. \emph{Evaluation protocol:} The eruption window (Dec~2024--Jan~2025) is fully held out from Stages~2/3 training, which uses only pre-eruption acquisitions. The 80/20 spatial split is maintained at evaluation: ROC/PR metrics are computed on the held-out 20\% spatial cells over the eruption window.

\subsection{Results}

\begin{figure}[t]
    \centering
    \includegraphics[width=\columnwidth]{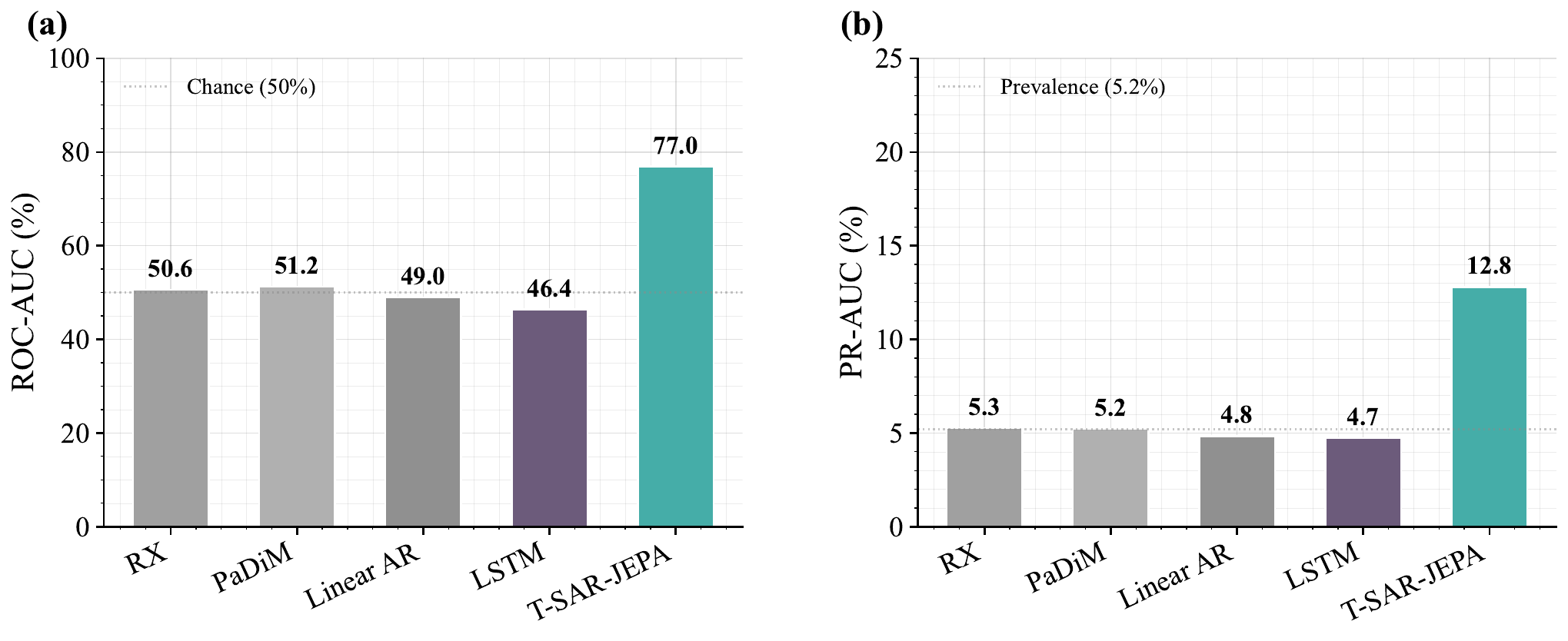}
    \vspace{-2mm}
    \caption{(a)~ROC-AUC and (b)~PR-AUC on Hawaii eruption-window evaluation (1{,}900 pairs, 99 positives at coherence $< 0.2$). T-SAR-JEPA (77.0\%) substantially outperforms all baselines (${\sim}50$\%).}
    \label{fig:roc}
\end{figure}

\textbf{ROC/PR Performance.} Table~\ref{tab:main_results} reports anomaly detection against coherence-drop pseudo-GT on 1{,}900 pairs from the Hawaii eruption window. T-SAR-JEPA achieves ROC-AUC of \textbf{77.0\%} and PR-AUC of \textbf{12.8\%}, substantially outperforming all baselines near chance (Fig.~\ref{fig:roc}). For each coherence pair $(t_a, t_b)$ with $t_a < t_b$, we assign the anomaly score $a_{t_b}$ of the later acquisition and label the pair as positive if mean grid-cell coherence falls below 0.2. This yields 1{,}900 scored pairs (99 positive, 5.2\% prevalence). PR-AUC is modest due to severe class imbalance (5.2\% prevalence) and label noise from coarse coherence pseudo-GT; ROC-AUC better reflects ranking quality. Hawaii is the only AOI with a known major event; LA and Pilbara are assessed via spatial coherence and geometry invariance below.

\begin{table}[t]
    \centering
    \caption{Anomaly detection vs.\ coherence-drop pseudo-GT (Hawaii, 1{,}900 pairs, 99 positives). All methods use the same frozen ViT features. Best in \textbf{bold}.}
    \label{tab:main_results}
    \small
    \begin{tabular}{lcc}
        \toprule
        \textbf{Method} & \textbf{ROC-AUC (\%)} & \textbf{PR-AUC (\%)} \\
        \midrule
        RX~\cite{reed1990rx}            & 50.6 & 5.3 \\
        PaDiM~\cite{defard2021padim}    & 51.2 & 5.2 \\
        Linear AR                         & 49.0 & 4.8 \\
        LSTM                              & 46.4 & 4.7 \\
        \textbf{T-SAR-JEPA (ours)}       & \textbf{77.0} & \textbf{12.8} \\
        \bottomrule
    \end{tabular}
\end{table}

\textbf{Spatial Coherence.} Table~\ref{tab:permutation} reports 4-connected neighbor agreement (fraction of cells matching all spatial neighbors' binary labels) at P80: 99.9\% for Hawaii and LA, 93.6\% for Pilbara, all with $p < 0.001$ (1{,}000-shuffle permutation test), exceeding null by $1.6{-}1.8{\times}$. High absolute values reflect the $10{\times}10$ grid where correlated events produce nearly uniform labels.

\begin{table}[t]
    \centering
    \caption{Spatial coherence: observed neighbor agreement vs.\ null distribution (1000 permutations). All $p < 0.001$.}
    \label{tab:permutation}
    \small
    \setlength{\tabcolsep}{3pt}
    \begin{tabular}{lccccc}
        \toprule
        \textbf{AOI} & \textbf{Thresh.} & \textbf{Observed} & \textbf{Null} & \textbf{Ratio} & \textbf{$p$} \\
        \midrule
        Hawaii  & P80 & 99.9\% & $54.8 \pm 1.0$\% & 1.82$\times$ & $<$0.001 \\
        LA      & P80 & 99.9\% & $54.8 \pm 1.3$\% & 1.82$\times$ & $<$0.001 \\
        Pilbara & P80 & 93.6\% & $58.1 \pm 1.4$\% & 1.61$\times$ & $<$0.001 \\
        \bottomrule
    \end{tabular}
\end{table}

\textbf{Geometry Invariance.} Table~\ref{tab:geometry} reports Spearman correlations between anomaly scores and satellite identity (all $|\rho| < 0.11$).

\begin{table}[t]
    \centering
    \caption{Geometry invariance: Spearman $\rho$ between anomaly score and satellite ID. All $|\rho| < 0.11$ (negligible effect size); small $p$-values reflect large $N$, not meaningful association.}
    \label{tab:geometry}
    \small
    \begin{tabular}{lccc}
        \toprule
        \textbf{AOI} & \textbf{$\rho$} & \textbf{$p$-value} & \textbf{Satellites} \\
        \midrule
        Hawaii   & 0.061 & $<$0.001 & C10, C13, C14 \\
        LA       & $-$0.102 & $<$0.001 & C07, C13, C14 \\
        Pilbara  & 0.044 & $<$0.001 & C09, C10, C14 \\
        \bottomrule
    \end{tabular}
\end{table}

\begin{figure}[t]
    \centering
    \includegraphics[width=\columnwidth]{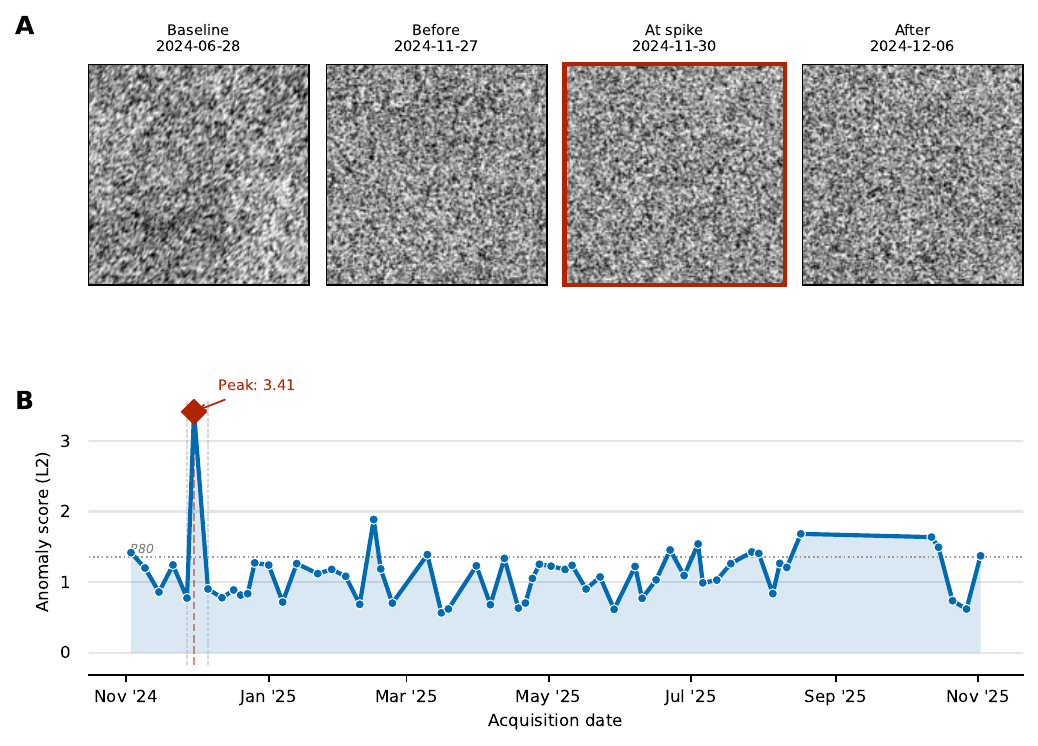}
    \vspace{-2mm}
    \caption{Kilauea case study. \textbf{(A)}~SAR patches at four dates: \emph{baseline}, \emph{before}, \emph{at spike} (red border), and \emph{after}. \textbf{(B)}~Anomaly score with P80 threshold (dotted) and peak (3.41) at eruption onset; flagged from amplitude alone.}
    \label{fig:qualitative}
\end{figure}

% ============================================================
\vspace{-1mm}
\section{Discussion and Conclusion}
% ============================================================
\vspace{-1mm}

\subsection{Ablation Studies}
\label{sec:ablations}

Table~\ref{tab:ablation} summarizes ablations. Sinusoidal encoding achieves the lowest validation loss; with only 300 temporal sequences, learnable encodings lack sufficient data, while fixed sinusoidal frequencies provide multi-scale periodicity without calibration. Performance peaks at $K{=}7$ (${\sim}$3 weeks at typical cadence); $K{=}9$ overfits and $K{=}3$ provides insufficient context.

\begin{table}[!tb]
    \centering
    \caption{Ablation results (validation loss $\times 10^{-3}$). \textbf{Top}: time encoding. \textbf{Middle}: context window $K$. \textbf{Bottom}: freezing strategy (Stage~3 adds 50 epochs).}
    \label{tab:ablation}
    \small
    \begin{tabular}{lc}
        \toprule
        \textbf{Time Encoding} & \textbf{Val Loss ($\times 10^{-3}$)} \\
        \midrule
        \textbf{Sinusoidal (ours)} & \textbf{2.70} \\
        Linear learnable      & 4.65 \\
        CTLPE~\cite{ctlpe2024} & 4.69 \\
        \midrule
        \textbf{Context $K$}  & \textbf{Val Loss ($\times 10^{-3}$)} \\
        \midrule
        $K{=}3$  & 3.42 \\
        $K{=}5$  & 2.47 \\
        $\mathbf{K{=}7}$  & \textbf{2.00} \\
        $K{=}9$  & 2.23 \\
        \midrule
        \textbf{Freezing}         & \textbf{Val Loss ($\times 10^{-3}$)} \\
        \midrule
        Frozen encoder            & 2.00 \\
        \textbf{Progressive (ours)} & \textbf{0.04} \\
        \bottomrule
    \end{tabular}
\end{table}

\textbf{Threshold Sweep.} Spatial coherence remains $>$91.7\% across P70--P90 for all AOIs, indicating robustness to threshold choice.

\subsection{Kilauea Eruption Case Study}

Fig.~\ref{fig:qualitative} shows the model's response to the December 2024 Kilauea eruption~\cite{usgs_kilauea}: the L2 prediction error on this grid cell peaks at 3.41 at eruption onset --- the only acquisition exceeding P80 --- driven by the visible amplitude texture shift in panel~A. InSAR coherence decorrelation at matching locations independently confirms the detection, despite the model never seeing coherence.

\begin{figure}[t]
    \centering
    \includegraphics[width=\columnwidth]{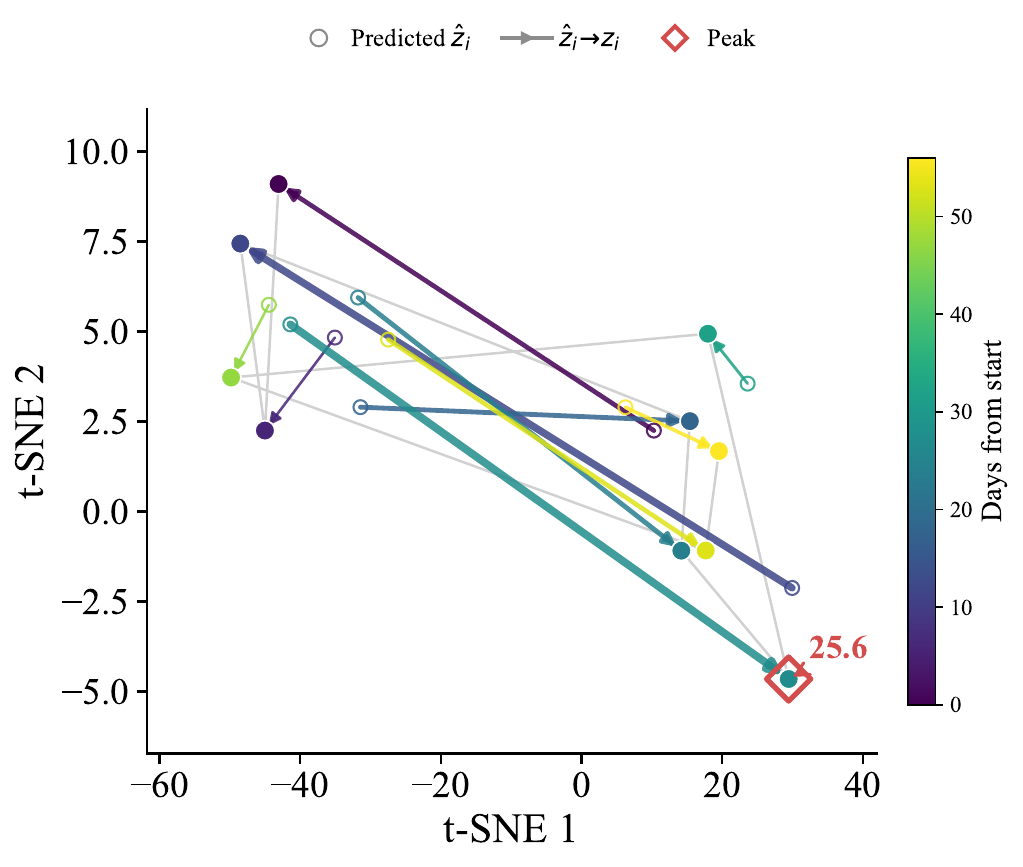}
    \vspace{-2mm}
    \caption{Latent state transitions across the eruption window (t-SNE, $\mathbb{R}^{768}{\to}2$D). Points: observed $z_i$ colored by relative day. Arrows: predicted $\hat{z}_i{\to}z_i$, thickness ${\propto}$ L2 error. Thin arrows = accurate prediction (stable surface); thick arrows = large prediction error (anomaly). Peak marked at 25.6.}
    \label{fig:latent_transitions}
\end{figure}

\textbf{Latent State Transitions.} Fig.~\ref{fig:latent_transitions} shows ten consecutive embeddings from the case-study grid cell spanning the eruption window, projected via t-SNE. Points are colored by relative acquisition day (viridis); each point trails an arrow from its predicted embedding $\hat{z}_i$ to the observed $z_i$, with thickness proportional to L2 prediction error. Pre-eruption acquisitions produce thin arrows (predictor tracks reality); at and around the eruption peak the arrows lengthen dramatically, directly visualising how the temporal transformer's inability to anticipate the surface change generates the anomaly signal.

\textbf{Calibration and Leakage.} Progressive unfreezing shifts score distributions (mean 0.453 vs.\ frozen 6.89); we use per-AOI P70--P90 thresholds for robustness. Stage~1 adapts on all AOIs using only the local masked reconstruction objective (no temporal labels or coherence), so information leakage is minimal; a leave-one-AOI-out protocol is planned.

\textbf{Limitations and Future Work.} Coherence pseudo-GT conflates physical change with temporal decorrelation, so ROC-AUC is a lower bound (threshold sweep over 0.15--0.35 planned). Event-level pseudo-GT exists only for Hawaii; LA and Pilbara rely on structural proxies. Our temporal baselines are simple; stronger architectures (TCNs, Transformers with matched unfreezing) are needed for fair ablation. Future work: bootstrap CIs for ROC/PR and formal annotations.

\textbf{Conclusion.} T-SAR-JEPA detects temporal anomalies in amplitude-only SAR via self-supervised latent prediction, without labels or interferometric processing. InSAR coherence validates ranking (ROC-AUC 77.0\% vs.\ ${\sim}50$\%), spatial structure ($>$93.6\% neighbor agreement, $p{<}0.001$), and geometry invariance ($|\rho|{<}0.11$) across three diverse AOIs.

\bibliographystyle{IEEEtran}
\bibliography{refs}

\end{document}